\documentclass[11pt]{article}
\usepackage[utf8]{inputenc}
\usepackage{amsfonts,fullpage}
\usepackage{amsgen,amsmath,amstext,amsbsy,amsopn,amssymb,subcaption}
\usepackage[dvips]{graphicx}
\usepackage{comment}
\usepackage{tabularx}

\newcolumntype{C}{>{\centering\arraybackslash}X}
\usepackage{graphicx}

\usepackage[shortlabels]{enumitem}
\usepackage{natbib}
\usepackage{booktabs}
\usepackage{amsthm}
\usepackage{hyperref}
\usepackage{blkarray}
\usepackage{algorithm}
\usepackage{algorithmic}
\usepackage{url}
\usepackage{longtable}
\usepackage{stmaryrd}
\usepackage{mwe}
\usepackage{mathtools}
\DeclareMathAlphabet\mathbfcal{OMS}{cmsy}{b}{n}

\makeatletter
\newcommand*{\rom}[1]{\expandafter\@slowromancap\romannumeral #1@}
\makeatother

\usepackage[dvipsnames]{xcolor}

\theoremstyle{plain}

\title{A Study of Data-driven Methods for Inventory Optimization}

\author{$\text{Lee Yeung Ping}^{\dag}, \text{Patrick Wong}^{\dag}, \text{Tan Cheng Han}^{\ddag}$\\
$\dag~$Hong Kong Baptist University\\
$\ddag~$Hong Kong University of Science and Technology
}
\date{}

\begin{document}

\maketitle

\begin{abstract}
This  paper  shows  a  comprehensive  analysis  of  three  algorithms  (Time Series,  Random Forest (RF) and Deep Reinforcement Learning) into three inventory models (the Lost Sales, Dual-Sourcing and Multi-Echelon Inventory Model). These methodologies are applied in the supermarket context. The main purpose is to analyse efficient methods for the data-driven. Their possibility, potential and current challenges are taken into consideration in this report. By comparing the results in each model, the effectiveness of each algorithm is evaluated based on several  key  performance  indicators,  including forecast accuracy, adaptability to market changes, and overall impact on inventory costs and customer satisfaction levels. The data visualization tools and statistical metrics are the indicators for the comparisons and show  some  obvious  trends  and  patterns  that  can  guide  decision-making  in  inventory management. These tools enable managers to not only track the performance of different algorithms  in  real-time  but  also  to  drill  down  into  specific  data  points to  understand the underlying  causes  of  inventory  fluctuations.  This  level  of  detail  is  crucial  for  pinpointing inefficiencies and areas for improvement within the supply chain. 
\end{abstract}

\section{Introduction}\label{sec:intro}

Successful inventory management is key to any retail business's success, particularly in supermarkets, where inventory plays an integral part in customer satisfaction and financial performance. Supermarkets face numerous complex challenges involving managing a wide range of products and perishable goods, rapidly shifting consumer tastes, and adapting traditional inventory methods less effectively, resulting in suboptimal operational outcomes \citep{powell2007approximate}. 
Time Series analysis and Random Forest (RF) algorithms have long been relied upon as tools for forecasting inventory needs using past sales data. While such approaches offer some predictability and control, they often prove insufficient in environments that are highly dynamic and unpredictable; real-time adaptation to new information quickly is thus required \citep{silver2016mastering}. 
The emergence of advanced analytics and machine learning technologies has opened new avenues for addressing complex challenges. One such technology, Deep Reinforcement Learning (DRL), specifically with Deep Networks (DQN), has emerged as an innovative 
solution. DRL marks a shift away from predictive analytics toward more flexible, responsive methodologies by combining principles such as reinforcement learning - making decisions   under uncertain conditions - with capabilities of deep neural networks, which help systems  learn the optimal actions through trial and error \citep{mnih2015human}. 

This report seeks to evaluate the implementation of DQN inventory management technology in supermarket inventory management and compare it with conventional approaches such as TS and RF. The comparative analysis is designed to illuminate both the advantages and disadvantages associated with each method, with specific emphasis being put on their ability to address challenges facing Hong Kong supermarkets. Evaluation criteria include handling lost sales, optimizing dual sourcing strategies, and the effectiveness of multi-echelon stock systems as important performance indicators. 

Supermarkets oversee an enormous variety of product lines with differing demand patterns and shelf lives, each one demanding precise attention from inventory managers. Due to 
perishability issues with some items, managing a vast assortment can be particularly 
complex; overstocking perishable items can incur substantial costs in terms of wastage also due to limited shelf space, while inadequate stock levels lead to decreased sales and 
unhappy customers, resulting in decreased customer loyalty that ultimately harms a supermarket's reputation and long-term customer relations. 

Conventional methods of analysis include Time Series and Random Forest models. 
Time Series Analysis and Random Forest algorithms have long been at the core of inventory forecasting. Time Series analysis allows us to utilize historical data over an extended period  in order to gain insights and predict consumer demand patterns. Random Forest, on the 
other hand, is an advanced machine-learning methodology that employs a collection of 
decision trees to generate predictions. It has become widely respected due to its exceptional precision, capacity to manage large datasets with multiple variables, and ability to offer invaluable insight into various predictors. 

Time Series analysis relies on the assumption that past patterns will continue in the future; however, this assumption often breaks down in an unpredictable retail market environment.   While Random Forest can provide powerful analysis capabilities, its algorithm often lacks the agility to adjust dynamically to any new changes or shifts that arise within its bounds. 

Deep Q-Networks have rapidly emerged in inventory management. By combining deep 
learning's resilience and Q-learning's decision-making ability into one approach, Deep 
Q-Networks provides a new and novel solution. Through iterative interactions and the 
incorporation of new data sets into iterative interactions between DQN systems and market fluctuations, their forecast accuracy increases, and they can adapt in real time as market     fluctuations occur. Deep Q-Networks are particularly helpful when applied to supermarket    environments where demand can quickly change due to factors like seasonal variations, 
promotions, or actions taken by competitors - particularly effective inventory management solutions.This research is designed to identify and examine specific objectives. The report will be organized logically. 

The primary goal of this report is to conduct an in-depth comparative analysis of DQN, TS,    and RF methods used for supermarket inventory management in Hong Kong. The aim of the research is to offer recommendations to practitioners and add to the academic discussion of  advanced machine learning techniques in real-world settings by exploring how these 
methodologies address issues such as lost sales, dual sourcing issues, and multi-echelon stocking problems. 

\section{Preliminaries}
After mentioning the background, we focus on describing the lost sales, dual-sourcing and multi-echelon inventory models of data-driven inventory management. 
Their main processes, data and applications in the supermarket context are introduced to demonstrate how well they perform. 

\subsection{Tested Inventory Models}
\subsubsection{Lost Sales Inventory Model} 
The lost sales inventory model is designed to diminish the total inventory costs (the holding,  ordering and shortages costs) in order to lessen the consequences of lost sales. Besides, this model also optimizes the order quantity in each replenishment cycle and determines the optimum reorder point for shrinking the risk of stockouts and holding costs. 
To delve into this inventory model, several crucial components must be considered to comprehend their effects on the performance of the lost sales inventory model. 
Just pointed out inventory costs before, holding, ordering and shortage costs are included in this component. To control the inventory levels, the fiscal impact is thoroughly influenced by  these costs. Holding costs indicate the expenditure by storing the unsold goods. Ordering and shortage costs include the spending on shipping and handling fees for the essential 
replenishment inventory and the revenue lost because of stockouts respectively. 
The demand which is another indispensable component is according to a discrete or 
continuous rate. If the demand during the sales period is not achieved by the inadequate  inventory supply, the lost sales would happen. In addition, If demand surpasses available stock and cannot be placed on backorder, the inventory level remains static when the 
system is depleted \citep{bijvank2011lost}. 
Therefore, the efficient inventory replenishment which is executed in determined intervals is covered by a well lead time. Consequently, the above content is summarized by decision 
variables order quantity (Q) and reorder point (s) and they indicate the quantity of ordering    inventory in each replenishment cycle and reorder threshold inventory level correspondingly. 

The lost sales inventory model requires comprehensive data utilization in the data-driven 
inventory management aspect \citep{chen2024learning}. Hinging on the historical sales data for the sluggish demand  format, diverse profitable alterations are the outstanding approach for grocery and hard lines \citep{fisher2010new}. Therefore, after understanding the habits of customers, some 
algorithms are used to predict future demand precisely by adjusting the order quantities and reorder points. 
Relying on the historical data, some insights into customer behaviour are shown for 
analyzing their buying habits in the stockout period. For instance, customers may wait for items to be restocked, and they may choose another brand to deal with the temporary 
product shortage. 
Besides, following and figuring out the items in each category helps study the trend and the inventory depletion. If the comprehensive analysis is done in the preceding period, the 
replenishment methods could be implemented well and could avoid leading to severe lost sales.

Another type of historical data is about lead time. Replenishment lead time and the length of product life directly affect the life cycle planning. There are three kinds of cases for diverse    sizes of lead time. “One and done” mode is operated by doing a single purchase and an 
unattainable replenishment is occurred if the lead time is larger than or equal to the product   life. An item is acquired a rapid replenishment after launch if the lead time is shorter than the product life several times. If the lead time is slightly lower than the product life, limited 
replenishments are feasible. An alternative is to make a single purchase from a supplier and distribute some portions of the purchase to a distribution center \citep{fisher2010new}. 
Therefore, utilizing historical data on lead times assesses the reliability of suppliers and manages lead time variability, which is crucial for setting realistic reorder points and 
minimizing the risk of stockouts. 
As a result, the lost sales inventory model could be applied in the supermarket context by changing reorder points and understanding the demand patterns. This model provides an outline for implementing different machine learning algorithms to analyze historical sales data, improving demand forecasting accuracy.

\subsubsection{Dual-Sourcing Inventory Model} 
The replenishment from two sources in an unreliable demand inventory environment belongs to the dual-sourcing inventory model. This model is always given by the assumption that one of the sources is cheap and is called a regular source. Another which is rapid is called express source. 
Furthermore, this model includes two time models which are divided into discrete and 
continuous. The former model considers the decision at specific and separated points in time, especially in regular intervals. In the prearranged intervals, the inventory decisions which are executed are consistent with the review periods. The latter model processes the unceasing inspection and approves the immediate decision-making. 
Several decision variables play an important role in the dual-sourcing inventory model. As 
mentioned in section 2.1.1, order quantities (Q) and reorder points (s) are included in this 
model but they are divided into the express source (QE  , SE) and regular source (QR  , SR). The order quantities are determined in each period according to the inventory regulation and the   present inventory levels. The reorder points represent the threshold to prompt the ordering     when the inventory level keeps decreasing below these points. The order up to levels (SE ,  SR ) represents the objective inventory quantity to which stock needs to be replenished when 
receiving orders from express and regular sources. To prevent stockouts due to demand and supply delays, safety stock levels (SS) indicate the quantity of inventory kept on hand to 
provide a buffer. 

For the dual-sourcing inventory model, the historical sales data is fruitful for the trend 
analysis. The seasonal fluctuations and growth patterns are identified by the sales trend over time. It is also useful for performance measurement by appraising the preceding results to 
make reasonable future ordering and stock level modifications. Besides, to utilize information technology-enabled sales and data from the market, it is viable to enhance the anticipation quality in the decision process. The new information which gives  a significant advance in demand forecast is in endogenous and exogenous forms \citep{cheaitou2014optimal}. In order to measure the stockout probability and lost sales, two variables which are set for the calculation are $D_1$  being the demand in the first lead time $L_1$ and $D_2$  being the demand in the second lead time $L_2$. Also, the assumption for the demand $D$ is to follow the normal distribution with mean $\mu$ and variance $\sigma_2$ \citep{fong2000analysis}. Thus, the demand data is used to model demand in the lead time period for 
computing stockout probabilities and potentially lost sales and for predicting future demand statistical models. 
The replenishment order amount is divided between two suppliers in a dual-sourcing 
inventory system, and orders are placed with each source simultaneously. The ordering 
process begins with the same trigger as for a single provider. The order would be divided if it is decided to restock \citep{fong2000analysis}. The operational decision-making 
process in a dual-sourcing system, entails using supplier data on lead times and stock levels to determine when to restock. 
Reorder points and order quantities need to be specially tailored to fit the distinctive sales 
trends and seasonal fluctuations in supermarket sales data in order to apply the dual-sourcing inventory model in a supermarket scenario. 

A dynamic retail environment needs careful calibration of safety stock levels and order-up-to  levels in order to handle the variety and speed of changing inventory demands. This ensures constant product availability and lowers the danger of stockouts. 

\subsubsection{The Multi-Echelon Inventory Model} 

The Multi-Echelon Inventory Model aims to optimize the inventory from some stages or levels 
in the supply chain and to curtail the total costs (holding costs, ordering costs and 
transportation costs) for keeping the apt service levels. When it comes to the restrictions and assumptions in this model, the lead times are either zero or constant and deterministic while  demands for every item arise at a consistent and deterministic rate. Assembly lines for 
intermediate goods follow precise, regular timetables, these presumptions could be true. 
Nevertheless, even with the most advanced forecasting systems, sales volumes in most contexts are susceptible to a significant degree of uncertainty and non-stationarity. 

Production and distribution lead times are frequently accompanied by comparable nonstationarities and uncertainty \citep{federgruen1993centralized}. 
Tactical coordination assures maximization of the overall performance by the 
decision-making in all echelons and within the supply chain, it improves efficiency and 
lowers costs by coordinating the goals and operations of different inventory holding points. To adapt to the specific characteristics of each echelon in some dynamic replenishment 
tactics, it maximizes the flow of commodities from upstream to downstream levels and guarantees that each echelon keeps an adequate amount of inventory on hand to satisfy demand without overstocking. 

To scrutinize this model, some components are taken into consideration. Echelons are the different layers where the inventory is occupied, such as warehouses and distribution 
centers. The echelons are regarded as the buffers to satisfy downstream demand efficiently.  Within each echelon, there are designated areas known as Decision Points where significant decisions about inventory levels, order quantities, and replenishment schedules are made. These choices are essential for computing the best inventory strategies that dynamically 
adjust to continuous shifts in supply and demand. 

The nodal enterprise of the layer k can supply goods, accessories and raw materials to the nodal enterprise of the layer k  +  1. However, no logistics between nodal enterprises in the same layer or in levels which are not contiguous do not happen.  Taking into account the interdependencies and relationships among diverse levels or strata in the supply chain. 
A centralized data repository that gathers and handles information from all echelons is a 
common characteristic of the Multi-Echelon Inventory Model. By utilizing aggregate and 
thorough data analysis to precisely estimate demands and efficiently manage inventory 
plans, this centralized method allows for complete visibility over the whole supply chain and improves decision-making. 

In order to optimize stock levels to satisfy steady customer demand while reducing 
expenses, the Multi-Echelon Inventory Model helps manage inventory across several levels, including individual storefronts and regional distribution facilities. Supermarkets diminish 
out-of-stock scenarios and excess inventory by employing a centralized data system to 
precisely estimate demand, make real-time inventory adjustments, and guarantee effective replenishment schedules. 

\subsection{Selected Algorithms}
\subsubsection{Time Series Method}
A Time Series model is a set of data points ordered in time, where time is the independent variable. These models are used to analyze and forecast the future \citep{mills2019applied}. It 
provides valuable insights into the underlying patterns, trends, and dependencies present in sequential data, making it a powerful tool for forecasting future trends and making informed decisions. 
In the field of inventory management, the Time Series model has been noticed to be an 
efficient forecasting method compared to some traditional economic systems. As \cite{chopra2001strategy} noted, A demand forecasting system that can reduce the inventory carrying cost and/or improve service levels can give manufacturing, distribution, and retail firms a significant competitive advantage. Therefore, \cite{hill2015forecasting} observed that Inventory 
managers need an economic model that can be used to estimate the future economic advantage of a Time Series forecasting system over a reorder point system. 

Time Series analysis in inventory management is mainly used to make forecasting   comparisons.\cite{liashenkoapplication} used Time Series analysis to proffer a 
methodological framework for the comparative evaluation of Time Series models, facilitating 
the judicious selection of the most optimal model to enhance forecasting accuracy. Their research conducts a comparative examination that includes traditional Time Series 
forecasting methods alongside some more advanced alternatives. 

An essential aspect of Time Series analysis is its abundance of advanced methods available for application and further utilization. Interrupted Time Series is an example of an advanced application of Time Series analysis. It is a strong design to use to estimate the effects of your product when randomization is not suitable or possible. As \cite{ejlerskov2018supermarket}) stated the controlled, interrupted Time Series approach used in their longitudinal analyses is one of the  strongest quasi-experimental designs available, it can easily show the strength of the 
variants of the Time Series model. 

In conclusion, Time Series analysis is a common predictive analytics model that is widely utilized in the field of inventory management. In addition to its inherent forecasting 
capabilities, it can be further enhanced through advanced applications, allowing for more accurate predictive models and providing a greater variety of options for comparison.

\subsubsection{Random Forests}

Breiman introduced the Random Forest (RF) algorithm in 2001 as an ensemble learning 
method. Researchers and practitioners have extensively studied and applied it in various 
domains, including inventory management. The RF algorithm constructs multiple decision 
trees at training time and outputs the class, that is, the mode of the classes or the mean 
prediction of the individual trees. Its robustness and accuracy in classification and regression tasks have made it famous. 
In inventory management, researchers have particularly praised RF for its predictive 
capabilities. \cite{moro2014data} noted the algorithm's effectiveness in forecasting product demand, a key component in managing inventory levels. RF handles high-dimensional datasets efficiently, providing reliable demand forecasts even with nonlinear relationships and interactions among variables, as \cite{genuer2010variable} observed. 

\cite{verikas2011mining} highlighted RF's versatility in processing various data types, including  categorical and continuous variables, without needing transformation. This feature benefits supermarkets that are dealing with a diverse product range. The algorithm's ability to 
manage missing values and maintain accuracy even with many missing data \citep{stekhoven2012missforest} is crucial in real-world inventory scenarios where data is often incomplete or sparse.

Beyond demand forecasting, \cite{sruthi2024improving} investigated RF's application in inventory classification, enhancing the inventory policy decision-making process. By categorizing items based on demand patterns and other characteristics, RF helps tailor specific inventory 
strategies to different product segments, potentially leading to significant cost savings and service level improvements. \cite{genuer2010variable}) also acknowledged RF's significance in feature selection, discussing its capacity to pinpoint the most critical predictors of demand among a potentially vast array of variables. This ability allows supermarkets to concentrate on the factors affecting inventory levels, optimizing their replenishment strategies. 
However, the RF algorithm has limitations. One challenge is determining the optimal number of trees and tree depth, which significantly affects performance, as \cite{oshiro2012many} 
pointed out. Although RF is known for its high accuracy, its computational intensity can be a drawback when quick decisions are necessary in fast-moving retail environments. 

In conclusion, the literature shows that RF is a potent tool for inventory management, 
offering robust forecasting and classification capabilities that can help optimize stock levels and enhance operational efficiency. Its implementation in supermarkets has contributed to  more accurate demand predictions and more effective inventory control, underscoring its relevance in the retail industry. 

\subsection{Deep Reinforcement Learning}
Deep Reinforcement Learning (DRL) is a sophisticated area of machine learning that 
combines the principles of reinforcement learning (RL) with the power of deep learning. 

\cite{arulkumaran2017deep} mentioned why deep learning was combined with reinforcement learning because they have a common goal of creating general-purpose artificial intelligence systems that can interact with and learn from their world. This fusion allows machines to 
make decisions and learn optimal behaviours in complex environments by experiencing and adapting through trial and error, much like humans learn from their own experiences. 
One of the main algorithms of DRL is Deep-Q-Network(DQN). \cite{mnih2016asynchronous} use recent advances in training deep neural networks to develop a novel artificial agent, that can learn 
successful policies directly from high-dimensional sensory inputs using end-to-end 
reinforcement learning. \cite{geevers2020deep} mentioned it is the first reinforcement learning 
algorithm that could directly use the images of the game as input and could be applied to numerous games. 
To implement the DRL algorithm into the field of inventory management, \cite{oroojlooyjadid2022deep} propose an enhanced DQN algorithm to play the beer game. The beer game is 
frequently employed in supply chain management courses to illustrate the bullwhip effect  and underscore the significance of coordinating supply chain activities. From applying the DRL algorithm to play the beer game, \cite{oroojlooyjadid2022deep} noticed that it obtains near-optimal solutions when playing alongside agents who follow a base-stock policy and perform much better than a base-stock policy when the other agents use a more realistic  model of ordering behaviour. Various deep RL algorithms have been further developed to handle inventory optimization problems in \cite{gijsbrechts2022can, madeka2022deep, alvo2023neural, liu2024reinforcement}

In summary, Deep Reinforcement Learning is a machine learning technique that is not only widely applied across various domains but also applicable in the field of inventory 
management. The Deep Q-Network Algorithm, in particular, has been demonstrated in 
previous studies to provide excellent analytical results in supply chain and inventory 
management. This underscores its potential for further advancement in predictive analytics within the inventory management domain by integrating machine learning technology. 

Based on the chosen models and algorithms, three algorithms are implemented in each model. After that, the details of implementations for the algorithms under the lost sales,  dual-sourcing and multi-echelon inventory model.

\section{Methodologies in Lost Sales inventory model}
\subsection{Time Series}
\subsubsection{Code Design}
The Time Series algorithm code is used the Prophet by Facebook for the model framework. Prophet is especially well-suited for real-world datasets, which frequently include gaps and anomalies because of its outstanding solving of missing data and outliers. More dependable forecasting is easily done by the distinguished ability of the Prophet.

The historical data on the estimated lost sales which are applied carries different proper columns, such as, ``Date'' and ``Estimated Lost sales values''. Next, good data preparation is necessary to ensure data is in the correct format. Thus, the ``Date'' column is converted into a ``Date Time'' format and sorted by date for sequential analysis. If the columns consist the missing data, they will be removed to guarantee the data quality and make the accuracy forecasting.

Furthermore, the code sets some configurations for the model. The changepoint and seasonality prior scale are respectively set to be 0.1 and 10.0 to recognize the trend with moderate flexibility and to fit the robustly seasonal variations for capturing the dynamics of lost sales.

In order to classify the annual patterns, the yearly seasonality is set up. The weekly seasonality enables us to observe the fluctuations in every week in lost sales to make some short-term inventory decisions. However, to concentrate on more general weekly and yearly trends, the daily seasonality is disabled in the code. These settings of seasonality aim for the lost sales anticipating in retail and other sectors with some sales cycles.

To talk about the forecast purpose, the next 180 days for predicting the future the lost sales are set. It brings some insights into the demand and supply discrepancies. It also allows the company's managers to observe the relevant insights and adjust the tactics. Some planning purchases, replenishments and logistics are previewed by the insights to prohibit lost sales.

In addition, training the Prophet model on our historical dataset leads to learning the specific patterns of lost sales to get authentic predictions. This procedure guarantees that the model is accurately calibrated to the historical data of a particular business environment.

The model validation is implied in the code because cross-validation is efficient in determining the predictive performance of the model by utilizing the historical data as the dataset. The initial size is set to be 180 and the period size is the range from the initial size dividing 10. The horizon size is ensured to be at least 30 but it does not exceed the remaining data length which is the difference between the datafrmae of prophet and initial size.

Other performance metrics such as Mean Absolute Percentage Error (MAPE) are the credible index to assess the model accuracy and tweaking parameters. Not only do they ensure that the forecasts are reliable and accurate, but they also lead to significant inventory decisions.

\subsubsection{Results}
As mentioned in the parameters and processes, some plots of forecast plots with change points, components MAPE and tables of performance metrics and forecast data are created.

\begin{figure}[!h]
    \centering
    \includegraphics[width=0.8\textwidth]{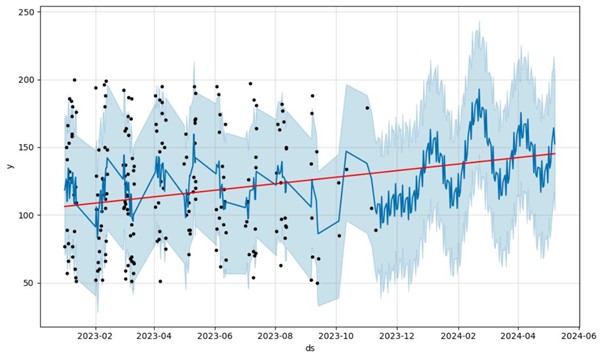}
    \caption*{Figure 1}
\end{figure}

This graph is the forecast plot which includes the historical data points, the forecast blue line and the uncertainty intervals shown by the shaded blue area. The trend component of the model is demonstrated as the red line. It indicates the general upward trend over time.

The x-axis which is labeled ``ds'' represents the time dimension, with dates ranging from early 2023 to mid-2024. The y-axis is labeled ``y'' which represents ``Estimated Lost Sales'' from the dataset. The y-axis values are the actual (the black dots) and predicted values (represented by the blue line and the blue-shaded uncertainty intervals) of the estimated lost sales over time.

\begin{figure}[!h]
    \centering
    \includegraphics[width=0.8\textwidth]{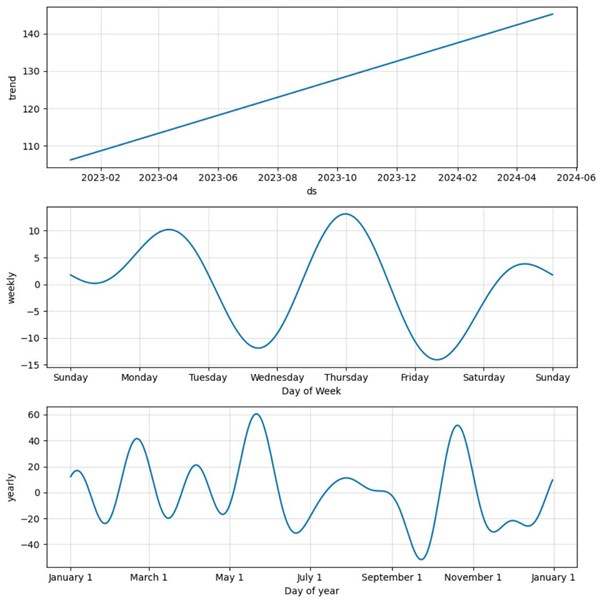}
    \caption*{Figure 2,3,4}
\end{figure}

The trend component of the Time Series data is displayed in Figure 2. It shows a distinct rising trajectory over time, indicating a general rise in the variable being monitored. The Time Series model date ``ds'' is used to indicate the x-axis, which runs from February 2023 to June 2024. The y-axis represents the magnitude of the trend component for ``Estimated Lost Sales''. The trend is that the model predicts that the amount of lost sales would increase with time in the absence of other factors, such as seasonality and vacations. This growing tendency may be the result of altering customer preferences or greater competition, two market dynamics that require attention.

The weekly seasonality is shown in Figure 3. and the weekly effect size on ``Estimated Lost Sales'' is displayed on the y-axis, while the x-axis indicates the days of the week. The changing blue line represents the variation in lost sales over the week. This pattern repeats every week and could assist firms in weekly strategic planning, such as staffing modifications or weekly promotions.

Figure 4 demonstrates the yearly seasonality. The x-axis represents the time of year (January to December) and the y-axis depicts the yearly impact on ``Estimated Lost Sales,''. Following the model understanding of how lost sales fluctuate over the year, the blue line waxes and wanes. Peaks might correspond with mass seasonal events that harm sales at specific times, while troughs might indicate periods of bellowing lost sales. This is essential for yearly strategic planning because it enables companies to anticipate sales downturns.

\begin{table}[!h]
    \centering
    \caption{}
    \begin{tabularx}{\textwidth}{|c|C|C|C|C|C|C|C|C|}
        \hline
          & horizon            & mse          & rmse       & mae        & mape      & mdape    & smape    & coverage \\ \hline
        0 & 0 days    08:31:00 & 838.091 687  & 28.94981 3 & 28.6212 44 & 0.2000 05 & 0.216712 & 0.205130 & 1.000000 \\ \hline
        1 & 1 days    09:51:00 & 1728.73 3182 & 41.57803 7 & 38.1397 86 & 0.5348 51 & 0.229909 & 0.376863 & 0.666667 \\ \hline
        2 & 2 days    02:19:00 & 1867.14 6027 & 43.21048 5 & 40.1829 29 & 0.6392 79 & 0.543195 & 0.450519 & 0.666667 \\ \hline
        3 & 19 days  23:35:00  & 2231.42 9308 & 47.23800 7 & 46.0053 49 & 0.6899 70 & 0.543195 & 0.483472 & 0.666667 \\ \hline
        4 & 20 days  04:45:00  & 3469.29 9976 & 58.90076 4 & 54.4073 36 & 0.6096 62 & 0.543195 & 0.450011 & 0.666667 \\ \hline
    \end{tabularx}
\end{table}

Table 1 demonstrates the cross-validation performance metrics of the Time Series (Prophet) forecasting model. The table provides a detailed insight into the forecast model performance at various time intervals into the future.

The time difference between the training data cut-off and the prediction made is represented by the column ``horizon.'' Throughout the cross-validation process, it shows the extent of the future predictions made by the model.

The average squared difference between the estimated and actual values is shown by ``mse'', which stands for Mean Squared Error. Estimator quality is gauged by MSE. Although values that are closer to zero are preferable, it is always non-negative. The square root of the mean of the squared errors is called Root Mean Squared Error presented in column ``rmse,''. Since it has the same units as the response variable, it is a useful indicator of how well the model predicts the response. It is also the most interpretable statistic.

The mean of the absolute values of the errors is the Mean Absolute Error, which is displayed in column ``mae.'' Without taking the direction of the errors into consideration, it determines the average size of the errors in a series of forecasts. The accuracy is expressed as a percentage of the error based on actual values, and the Mean Absolute Percentage mistake, which is displayed in column ``mape,'' is helpful for comprehending the prediction errors.

The Median Absolute Percentage Error is shown in the column ``mdape''. This is comparable to MAPE but it utilizes the median of the percentage errors rather than the mean, providing it a robust measure against outlier errors.

Symmetric Mean Absolute Percentage which is the alternative for MAPE is an error in column ``smape'' prevents that some zeros in the actual numbers from rendering MAPE ambiguous. It is robust to scale and provides an illusion of error proportionality by normalizing the errors with respect to both anticipated and actual values.

The ratio of actual values that fall in the prediction interval is shown by the ``coverage'' metric. A larger coverage percentage means that the predicted interval covers more actual values.

\begin{figure}[!h]
    \centering
    \includegraphics[width=0.8\textwidth]{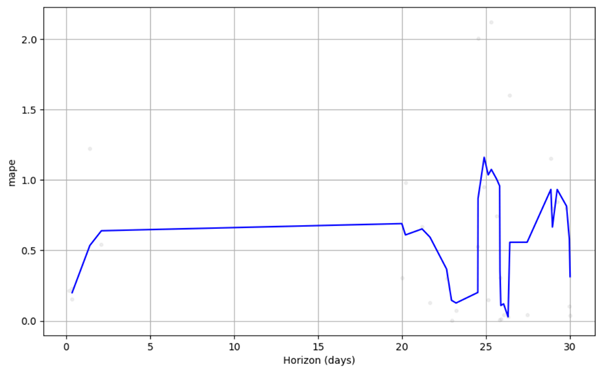}
    \caption*{Figure 5}
\end{figure}

Figure 5 is a plot of the Mean Absolute Percentage Error (MAPE) over the forecast horizon for a Time Series (Prophet) model.

The number of days into the future that the forecast is made is displayed by the graph's x-axis ``Horizon - days.'' For the y-axis ``mape'', means the Mean Absolute Percentage Error is the general measurement to test the accuracy of the prediction. By presenting the errors as a percentage, MAPE offers ideas of the magnitude of the error in comparison to the actual data. Forecasts with lower MAPE values are more accurate and those with greater MAPE values are less accurate. Therefore, the blue line is used to show the computed MAPE for each day in the forecast horizon.

To look at the spikes, the graph shows a sharp spike in 25-day. It is the day with
considerable fluctuation in the forecast inaccuracy. When the MAPE spikes, it indicates that the prediction is not as accurate as it usually is on a given day.

\begin{table}[!h]
    \centering
    \caption{}
    \begin{tabularx}{\textwidth}{|c|C|C|C|C|}
        \hline
            & ds                  & yhat       & yhat\_lower & yhat\_upper \\ \hline
        385 & 2024-05-03 20:30:00 & 138.860122 & 88.612856   & 186.948861  \\ \hline
        386 & 2024-05-04 20:30:00 & 152.461648 & 99.492135   & 206.250222  \\ \hline
        387 & 2024-05-05 20:30:00 & 158.472357 & 105.382582  & 206.866310  \\ \hline
        388 & 2024-05-06 20:30:00 & 164.228520 & 117.224742  & 216.999862  \\ \hline
        389 & 2024-05-07 20:30:00 & 152.412924 & 103.469020  & 200.705819  \\ \hline
    \end{tabularx}
\end{table}

Table 2 provides small parts of the output from the Time Series forecasting model in the future dates. ``ds'' includes the date and time that the predictions are made and ``yhat'' indicates the predicted value in the model and it is a great analysis for the target variable on specific date and time.

The lower bound of the forecast uncertainty range is given by ``yhat\_lower''. Given a specific confidence level, it shows the lower bound of the range within which the actual number is anticipated to fall. On the other hand, the upper bound of the forecast uncertainty range is displayed by the ``yhat\_upper''. It represents the upper bound of the range within which the actual value is anticipated to occur. Thus, the wider the range between ``yhat\_lower'' and ``yhat\_upper'', the greater the uncertainty occurred.

\subsection{Random Forest}
\subsubsection{Code Design}
The implement the Random Forest algorithm, the dataset CSV file is required and all the necessary libraries are imported before the following procedures. Then, it is crucial to convert the ``Replenishment Date'' into library ``pandas'' data format as it is more accessible to extract the year, month and day in each data. By doing this, the seasonal trends that have an impact on sales are captured. It is possible to spot patterns or seasonal trends influencing inventory availability. Therefore, lost sales by analyzing the timing of replenishments and sales. In inventory databases, where missing information distorts perceptions of sales, stock levels, and replenishment requirements, handling missing data is operated consequently.

In addition, the interaction term which is created is calculated with ``Quantity Sold\_x'' and ``Price'' to figure out some relationship between the number of sold and the price. High-value products are frequently sold out resulting in greater lost sales. The interaction assists in recognizing the revenue consequences of stock-outs.

For the Random Forest, ``Estimated Lost Sales'' in the dataset is selected to be the target and other columns such as ``Supplier ID Encoded'', ``Item ID Encoded'', ``Shelf-life'' and ``Replenishment Year'' are the features. Then, ``train\_test\_split'' from the ``scikit-learn'' is used to divide the dataset into training and testing sets. By testing the model using hypothetical data, this approach helps guarantee the model's generalizability and integrity. Using ``StandardScaler,'' features are normalized. During training, normalization offers a faster convergence rate. In order to ensure optimal performance of machine learning models, all features are on the same scale.

To manage the non-linear relationships and the feature interactions effectively, Random Forest Regressor and Gradient Boosting Regressor are selected to interact between distinct kinds of items and sales. ``RandomizedSearchCV'' is implied in the code for the hyperparameter tuning to optimize the settings. For instance, ``n\_estimators'' which is the number of trees is applicable for both RandomForest and Gradient Boosting. The maximum depth of each tree is displayed in ``max\_depth'' to avert the model from being too complex and overfitting the data. ``min\_samples\_leaf'' defines the minimal amount of samples that are required at a leaf node, while ``min\_samples\_split'' gives the minimum number of samples needed to split a leaf node.

Training each model on the scaled training dataset and utilizing the best hyperparameters in ``RandomizedSearchCV'' ensure that the model anticipates lost sales more accurately. MSE, MAE and $R^2$ scores of Random Forest Regressor and Gradient Boosting Regressor are expected to be shown. MSE and MAE indicate the average error margin and $R^2$represent the proportion of variance in the dependent variable which is predictable from the independent variables. These metrics help quantify the model performance based on the business impacts.

For further results, some learning curves are established to demonstrate the evolution of the validation and training scores with the addition of more data. This is beneficial in diagnosing issues such as underfitting where both scores are low or overfitting where the training score is high but the validation score is low.

There is a histogram of residuals to display differences between the observed and predicted values to find out the patterns with the frequency. Additionally, a quantile-quantile (Q-Q) plot helps to verify if the residuals are normally distributed, which is an assumption of many regression models.

\subsubsection{Results}

\begin{table}[!h]
    \centering
    \caption{}
    \begin{tabularx}{\textwidth}{|c|C|C|}
        \hline
              & Random Forest       & Gradient Boosting   \\ \hline
        MSE   & 0.505754022022078   & 0.39833127011749325 \\ \hline
        MAE   & 0.47562775427441145 & 0.37428327437660314 \\ \hline
        R$^2$ & 0.9998848826727091  & 0.9999093337290548  \\ \hline
    \end{tabularx}
\end{table}

Table 3 is one of the results of implementing the Random Forest algorithm in the lost sales inventory model.

As mentioned in the effect of MSE in section 3.2.1, a lower MSE illustrates a better fit of the model with the data. To compare the MSE of Random Forest and Gradient Boosting, the latter value performs slightly better to fit the data.

For the MAE, it is the average over the test sample of the absolute differences between the forecast and actual observation and all individual differences have the same weight. The result of MAE in Table 3 shows that Gradient Boosting shows a lower MAE indicating its predictions are closer to the actual values.

$R^2$ score gives a measure of the quality offit and how well the model is expected to predict unseen data. A 1.0 $R^2$ score contributes to a perfect prediction. $R^2$ score of both models which are close to 1 performs well for utilizing the chosen features to describe the variance in the target variable.

\begin{figure}[!h]
    \centering
    \includegraphics[width=0.8\textwidth]{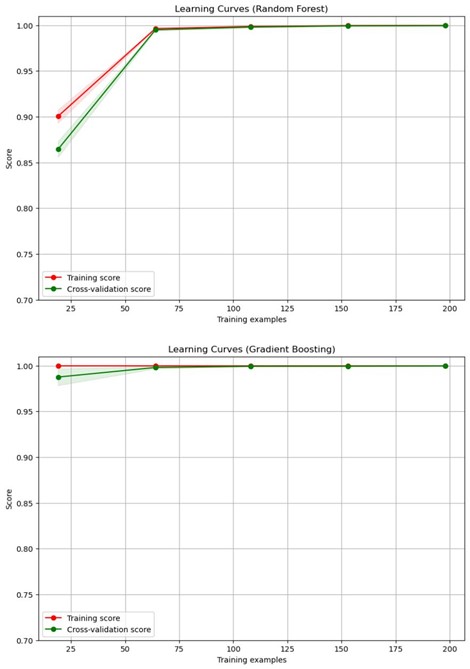}
    \caption*{Figure 6,7}
\end{figure}

In Figure 6, the training score is shown as a red line and begins at 0.9 scores in the training curve in Random Forest. It keeps rising close to a 1.0 score in training examples between 50 and 75. Since the model added more training examples, the score increases continuously and reaches to 1.0 score in in training examples between 100 and 125. As mentioned before, if the $R^2$ score is 1, it performs well for using the selected features to reflect the variance in the target variable. The green line is the cross-validation score and boosting continuously. The model performs better when the number of data is enhanced.

The result in these two figures reveals that the model does not emerge overfitting issue.

In Figure 7, the red line also represents the training score in the Gradient Boosting learning curve. It reaches a 1.0 score at the beginning, and even more data is added. Similarly, the green line is also the cross-validation score which starts from a high score. After raising the training data in the model, the score reaches 1.0 which same as the training score.

\begin{figure}[!h]
    \centering
    \includegraphics[width=0.8\textwidth]{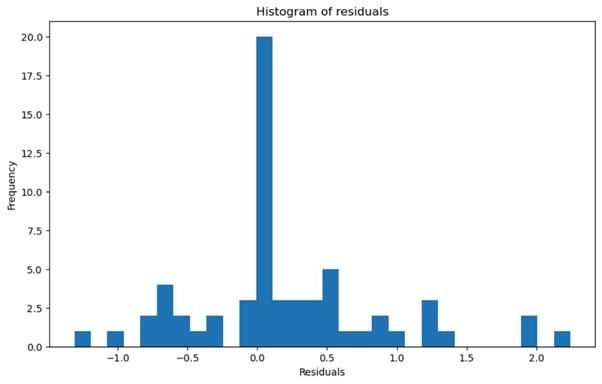}
    \caption*{Figure 8}
\end{figure}

The histogram of residuals is shown in Figure 8 with the frequency of residuals. By observing the residual data, most of them are concentrated in the centre close to 0.0 approximately. Therefore, there is no obvious and systematic bias in the model and the result of the model is acceptable.

Nevertheless, the distribution of the residuals in Figure 8 is not perfectly symmetric. The residuals get the highest frequency with 0.0 but the positive residuals are slightly more than the negative residuals. Some residuals not being 0.0 indicates that some outliers and noise emerge in the data to lead to an inaccurate prediction.

\begin{figure}[!h]
    \centering
    \includegraphics[width=0.8\textwidth]{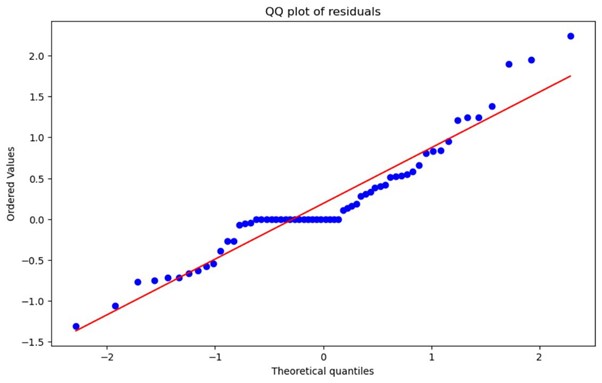}
    \caption*{Figure 9}
\end{figure}

Figure 9 shows a Q-Q plot of residuals to determine the credibility of data in some theoretical distributions with the x-axis ``Theoretical quantiles'' and y-axis ``Ordered Values''. The former value reveals the theoretical quantiles from the normal distribution. If the data is normally distributed, the value shows that the specific point is at the percentage of value. The latter value shows the ordered residuals from the dataset. The ordered residuals which are sorted by ascending order are the discrepancy between observed and forecasted values from the model. The red line represents that $y = x$. Therefore, the points lie on the red line if they follow the normal distribution.

In Figure 9, both sides' tails of the points at the end of the plot being slightly far from the red line show that there are heavier tails than the normal distribution in the residuals. In addition, there are more outliers than the normal distribution in the data.

\subsection{Deep Reinforcement Learning (DRL)}
\subsubsection{Code Design}
Before designing the main part of the DRL algorithm, the dataset file is adjusted and cleaned. The ``Date'' and ``Replenishment Date'' columns are converted to pandas library datetime format to operate the date-related functions. For the missing numeric values, the data is filled with the median of each respective column. ``Season\_winter'', ``Category\_Meat'' and ``Category\_Produce'' are the selected columns for the category, the missing values in those columns are filled with the mode of each column to guarantee the data is consistent.

When it comes to the main parts of the algorithm, they are formed by three components. The first part is the ``LostSalesInventoryEnv'' class which simulates the inventory management circumstance.

The cleaned dataset which is loaded includes some important columns in the constructor method ``def init (self, dataset)'', such as ``Estimated Lost Sales'', ``Estimated Demand'' and ``Lead Time''. The initial state is set to zero for starting the simulation step at the first row. Then, the action spaces are allowed with three possible actions. For example, zero represents the order stock, one indicates no actions and two illustrate diminishing prices. For the observation space, there are three continuous variables with 0 as the minimum value and no upper limit.

To define the cost parameters, the order cost per unit when ordering new stock is supposed to be 10 and the daily cost of holding inventory is 1. Furthermore, the stockout cost for each unit is 50 and the price reduction cost is 5.

After that, ``def reset(self)'' is designed to reset the environment to initiate the state and extract the ``Quantity Sold\_x'', ``Estimated Lost Sales'' and ``Days Until Replenishment''.

Next, the step function ``def step(self, action)'' is created to execute the action. The state and procedures alter according to the action. The reward value is set to 0 at the beginning.

If the ``order stock'' action is taken, the order quantity and the estimated lost sales would be assumed as 50 units and 0 respectively. The ``days\_untils\_replenishment'' is changed based on the ``Lead Time'' from the dataset and the reward is shrunk by the multiple of the order cost and the quantity ordered of product per unit. The reward is reduced by the product of the per-unit order cost and the quantity ordered. If the action equals 1, the state remains unchanged. In addition, if the ``Reduce Prices'' action is operated, the sales price would be decreased by 10 percent at maximum and the number of sales would also be raised by the percentage of the current quantity sold. The reward value is diminished by the price reduction cost 5.

To increase the current step value for the simulated progress by 1, the step would return to the starting point if the step comes to the end.

After processing entire steps, the reward value is able to be computed. The revenue is the multiple of the product of quantity sold and the current price per unit and the holding cost is calculated by multiplying the hypothetical hold cost per unit and by the fraction of the month until the next replenishment. The estimated lost sales cost is multiplied by the current price per unit to determine the lost sales cost. Moreover, the reward is lowered by a defined stockout cost if the quantity sold is less than the total of the estimated lost sales and the current estimated demand.

Deep Q-Network (DQN) agent part is also important to operate the learning process, balances exploration with exploitation, and utilizes past experiences through a neural network.

In order to operate the learning processes, harmonize exploration with exploitation and use the historical experiences through the neural network, the ``DQNAgent'' is established for making an effective result.

In the constructor of a class ``def init (self, state\_size, action\_size)'', the state and action sizes specify the input state vector dimension and the number of possible actions made by the agent correspondingly. The maximum value of 5000 is set by the ``self.memory'' to save the experience tuples. The discount rate and exploration rate are determined as 0.9 and 1.0 respectively. In addition, the epsilon is shrunk by multiplying with ``epsilon\_decay'' (= 0. 995) and stops at ``epsilon\_min'' (= 0. 01). The learning rate is set to 0.001 to identify the size of the step operated in the parameter space.

To initialize the model, the model is constructed by ``def \_build\_model(self)'' to execute the neural network configuration. There are two hidden layers with 64 neurons and ``relu'' activation contains the 0.1 rates of dropout layers. Utilizing the Adam for the optimizer is set with a certain learning rate and clipping the norm of gradients to be 1.0. Mean Squared Error (MSE) is used in the loss function to train the model to determine the squared difference between the forecast Q-values and the target Q-values.

To select the action, the ``def act(self, state)'' is established. If the exploration rate is larger than the random number, the random action will be selected. Otherwise, the agent would utilize the trained model to anticipate the most rewarding action for the current state.

Additionally, to replay the experience, a minimum batch size is formed by the random sample batch. The targeted Q-value is calculated by 2 circumstances for each experience in the minibatch. If it is not the end of the episode, the target is updated to be the summation of the reward, the discount rate and the maximum predicted reward value for the next state determined by the agent model. Differently, the target is identical to the reward. For the last step, the model and epsilon are updated. The neural network is trained by the Q-values for the batch with the single epoch to update the model. The epsilon decreases by multiplying the number of ``epsilon\_decay'' (= 0. 995) if the number of epsilon is greater than ``epsilon\_min'' (= 0. 01). The model loads the weights into the model from a file and saves the current weights of the model to a file.

To train the Deep Q-Network (DQN) agent in the environment, ``if name == " main "'' is set to ensure the script runs as the main program. At the beginning, the code initializes the environment from the mentioned class ``LostSalesInventoryEnv'' and the relevant dataset. The state size of the observation space from the environment is extracted to clarify the number of features obtained by the agent at each step. Similarly, the action size is determined by the environment to decide the number of possible actions chosen by the agent. ``agent = DQNAgent(state\_size, action\_size)'' is designed to execute the DQN algorithm.

For the training structure, the episode size is set to 100 for doing the training by the agent. The episode illustrates the comprehensive sequence from the beginning to the end. To train the rural network, the batch size is set to 10 for updating the network by the agent after processing and learning 10 samples. To store the rewards and epsilon, the metrics of their lists are created.

The progress bar is constructed by ``tqdm'' to track the training episode and processing episode and demonstrate the total estimated training time. To obtain the initial state at the beginning of each episode, the environment is reset and the state is reshaped to be suitable for the input shape. The total reward equals 0 in the initialization and It is accumulated by the rewards in each episode.

In the training step loop, the agent selects the action based on the current state and the environment executes the action by returning the new state and the reward from the action. To monitor the cumulative reward in the episode, the reward is summed into ``total\_reward''. Then, the state is updated and rephrased and the agent uses the ``remember'' approach to save the state transition. After that, there are two conditions to be checked. If the episode reaches the terminal state, a message is printed to reveal the current episode number, the number of time steps taken in this episode, and the current value of ``epsilon''. What is more, if the memory of the agent is larger than the ``batch\_size'', the agent has to review the batch of past experiences.

Lastly, ``rewards.append(total\_reward)'' represents the list storing the cumulative reward in the current episode and ``epsilons.append(agent.epsilon)'' indicates saving the current exploration rate for the episode into the epsilons list.

\subsubsection{Result}

\begin{figure}[!h]
    \centering
    \includegraphics[width=0.8\textwidth]{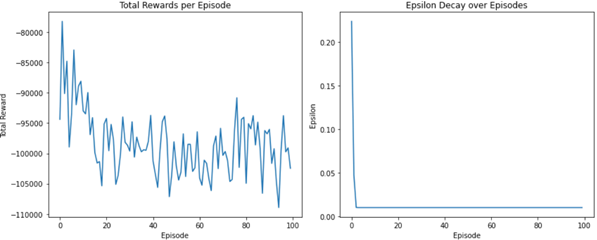}
    \caption*{Figure 10,11}
\end{figure}

To view Figure 10, the x-axis ``Episode'' illustrates the number of episodes in the training procedure and the y-axis indicates the accumulative reward from the agent. The blue line shows the fluctuation in the variability of the agent performance from one episode to the next.

At the beginning, the figure shows a shrink in the total reward performance. The value of the total reward mostly is between -9000 and -105000 and there is a plateau state beginning from episode 20 approximately.

For Figure 11, the graph shows the epsilon decay over episodes with the x-axis ``Episode'' and y-axis ``Epsilon''. The former axis represents the number of episodes identical to Figure 10. The latter axis illustrates the value of epsilon. The higher epsilon expresses the exploration which tries the new actions and the lower epsilon exploitation which chooses the best action.

The value of epsilon occurs a sharp diminishing from the first episode and it remains unchanged at the minimum value of epsilon (= 0. 01). The agent turns the exploration towards exploitation and mostly relies on the learned policy. If the rewards do not improve over time, this could indicate that the agent may not be learning an optimal policy or that the exploration rate is decreasing too quickly.

\subsection{Comparison in Lost Sales Inventory Model}

To consider the supermarket context in the lost sales inventory model, demand prediction, inventory management and lost sales are the possible consideration aspects.

To consider the demand prediction aspect, the results in the Time Series algorithm show that the error rises along the horizon length. This is capable of anticipating the short-term demand effectively. However, for the machine learning models (Random Forest and Gradient Boosting), the results from these two algorithms illustrate the high accuracy with the low MSE values and high $R^2$ scores. It contributes to the outstanding performance of the demand forecast demand and more accurate inventory planning. Random Forest and Gradient Boosting are better applied in this aspect in the lost sales inventory model since the fluctuating demand occurs in the supermarket context because of the seasonality, promotions and events. Thus, the Random Forest and Gradient Boosting are capable of optimizing the stock, shrinking the overstock and understock circumstances.

In inventory management, the accuracy in the Time Series shrinks over time so the Time Series algorithm is not the most suitable for long-term inventory management. However, the results of Random Forest and Gradient Boosting demonstrate they transcend for providing the explicit replenishment times to control inventory management well. The learning curve in the Deep Q-Network result does not show improvement and indicates that the DQN is not successful in learning the optimal inventory management policy. The perishable products and high holding and stockout costs are the main factors for inventory management in the supermarket context. Therefore, the most suitable algorithm for the above features is Random Forest ( and Gradient Boosting) as they own the high accuracy and the ability to change the model complexity.

To diminish the lost sales, the Time Series algorithm is able to assist in anticipating the demand peaks to avoid lost sales. Besides this, the effectiveness decreases over time and the lost sales in the long term potentially emerge. Although the real-world situation may not exactly match the training model, for the Random Forest (and Gradient Boosting) algorithm, the results in these algorithms represent the low error rates to be able to match the inventory levels with the demand.

\subsection{Discussion}

\begin{figure}[!h]
    \centering
    \includegraphics[width=0.8\textwidth]{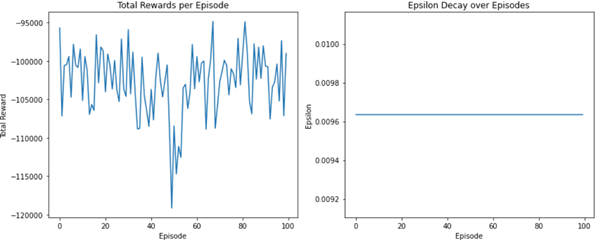}
    \caption*{Figure 12,13}
\end{figure}

After the initial results shown in the previous section, some parameters are changed to find a better performance in the DRL algorithm. The ``epsilon\_decay'' is altered from 0.995 to 0.885 and the learning rate boosts from 0.001 to 0.005. There are two hidden layers in the network architecture with 64, then increasing to 128 neurons and the dropout rate is also mounted to 0.3.

To execute the code with the adjusted code, the graphs is shown above. Figure 12 representing the total rewards per episode still reveals fluctuations. However, it is slightly more smoother than Figure 10.

After changing some parameters, the agent is able to learn in a more stable policy but there is no obvious upward trend for the total rewards. Figure 13 also indicates the exploration rate being more stable after shrinking the ``epsilon\_decay'' value. The agent is able to explore more over a large number of episodes before moving to exploitation.

The variance in rewards slightly diminishes and the agent performs a more consistent result after parameters tuning. Because of the slower epsilon decay shown in Figure 13, the agent is capable of exploring the environment for a longer period, it leads to better policies in a more complex environment. The smoother curve shown in the graph illustrates the more stable learning process.

\section{Methodologies in Dual-sourcing inventory model}
\subsection{Time Series}
\subsubsection{Code Design}
Most parts of the Time Series algorithm code in the dual-sourcing inventory model are the same as the code mentioned in section 3.1.1. However, to construct the specific code for the dual-sourcing inventory model, some adjustments are implied based on the code described in section 3.1.1.

First, the external regressor ``Promotion Type\_Discount'' in the feature engineering, which are filled for the missing values by using ``fillna(0)''. This adjustment is capable of improving the prediction accuracy by computing the new variability. After adding the external regressors, the regressors of the above column and ``Quantity Replenished'' are added to the Prophet model. The future values for all regressors are included in the dataframe ``future''.

To select the relevant columns for the Prophet model for the seasonality, the ``changepoint\_prior\_scale'' is slightly diminished to 0.05 to contribute to the less trend flexibility and the smoother trend alteration over time. In addition, the period is set to establish 3 folds within the initial training period and the difference between the maximum and minimum date indicates the number of total days. The horizon size is half of the number of total days.

In the last part of the algorithm, the decision systems integration is created. ``iterrows()'' is the approach to loop in dataframe rows and each iteration accesses the single row of dataframe ``forecast''. If the predicted value ``yhat'' from the forecasting model is larger than 100, a required action of the message for the predicted value ``row['yhat']'' and the corresponding date ``row['ds']'' is printed. Otherwise, a message is printed for no action required on the specific date and the anticipated number of units.

\subsubsection{Result}

\begin{figure}[!h]
    \centering
    \includegraphics[width=0.8\textwidth]{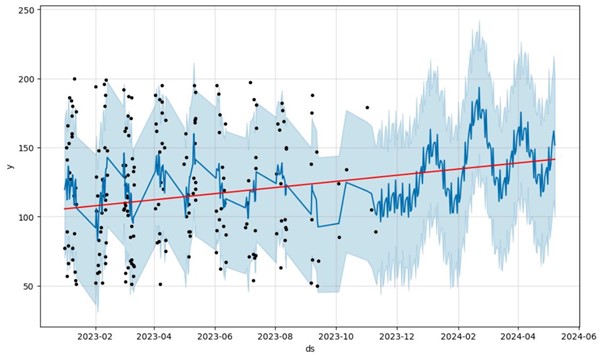}
    \caption*{Figure 14}
\end{figure}

Figure 14 shows extreme patterns with Figure 1 in section 3.1.2. Since the two algorithms apply in the same dataset with identical trends and seasonality. However, the predicted value represented by the blue line changes slightly from ``2023-10'' to ``2023-12''. Besides this alteration, all graphic elements, and the x\&y-axis remain unchanged.

\begin{figure}[!h]
    \centering
    \includegraphics[width=0.8\textwidth]{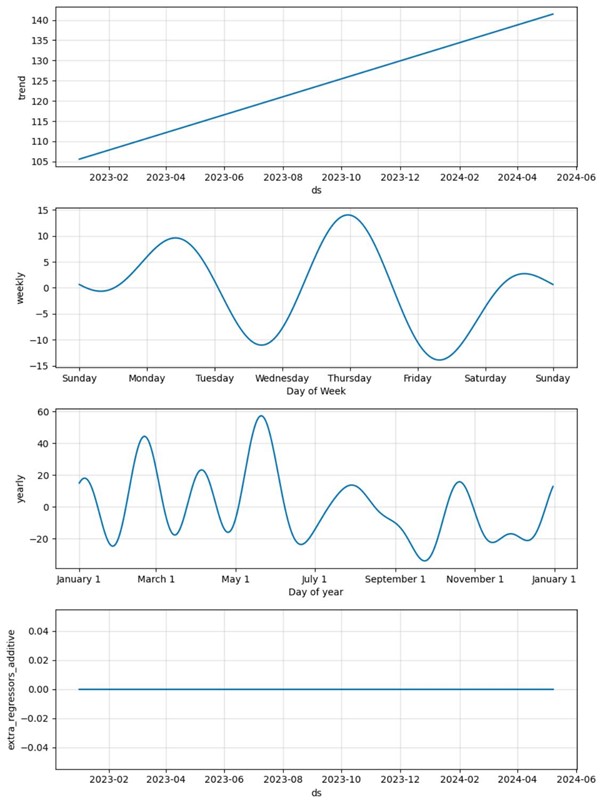}
    \caption*{Figure 15,16,17,18}
\end{figure}

Figure 15 also shows the trend of Time Series data. The elements represent identical meanings as the Figure 2. Nevertheless, the value in Figure 15 alters slightly and it is observed clearly on each date in the x-axis. To describe the difference, the slope of the blue line in the figure decreases and is smaller.

Figure 16 demonstrates an extremely akin shape and pattern to the figure in section 3.1.2. but the y-axis changes to ``Quantity Replenished'' rather than ``Estimated Lost Sales''. The x-axis is still the ``Day of Week''.

To express the meaning of figure 17, the time of year is illustrated in the x-axis while the y-axis alters to ``Quantity Replenished''. The value of ``Quantity Replenished'' suddenly shrinks in ``August 1'' approximately and the peak is half of the figure 4 in nearly ``September 1''.

In Figure 18, the graph illustrates the additional regressors added to the model to compute other factors except the trend and seasonality. The x-axis represents the date and the y-axis indicates the magnitude of factors that impact the forecast value. The blue line remains at 0 revealing the no effects from the external regressors.

\begin{table}[!h]
    \centering
    \caption{}
    \begin{tabularx}{\textwidth}{|c|C|C|C|C|C|C|C|C|}
        \hline
          & horizon           & mse          & rmse       & mae        & mape      & mdape     & smape     & coverage \\ \hline
        0 & 1 days   21:02:00 & 4485.22 1921 & 66.9717 99 & 51.5762 59 & 0.75746 5 & 0.44756 8 & 0.43117 9 & 0.50     \\ \hline
        1 & 2 days   00:58:00 & 4395.19 0177 & 66.2962 30 & 48.4663 60 & 0.73095 2 & 0.39454 2 & 0.40752 8 & 0.50     \\ \hline
        2 & 2 days   09:51:00 & 4361.13 5650 & 66.0388 95 & 48.3116 08 & 0.59801 3 & 0.39454 2 & 0.37159 4 & 0.50     \\ \hline
        3 & 2 days   19:46:00 & 4351.47 6017 & 65.9657 18 & 48.2453 51 & 0.66090 4 & 0.52032 5 & 0.40217 3 & 0.50     \\ \hline
        4 & 3 days   01:31:00 & 6428.92 4451 & 80.1805 74 & 70.4806 47 & 1.07985 4 & 1.29344 0 & 0.62788 4 & 0.25     \\ \hline
    \end{tabularx}
\end{table}

Table 4 also shows the cross-validation performance metrics and gives the predicted performance for future time intervals. The ``horizon'' column is half of the total days and demonstrates the anticipations in the future.

\begin{figure}[!h]
    \centering
    \includegraphics[width=0.8\textwidth]{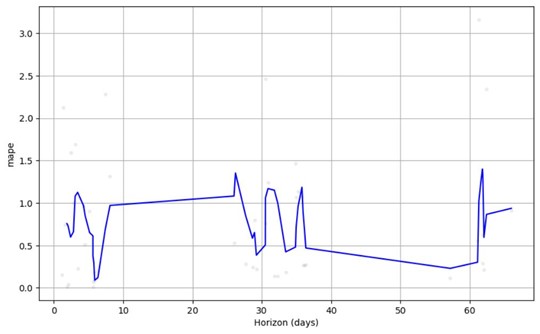}
    \caption*{Figure 19}
\end{figure}

In Figure 19, the variability shown in the above graph reveals that the performance of the model is inconsistent in the entire forecast horizon.

The high values of MAPE also show that the model was off by over 50\% or more of actual values in nearly half of the predictions. For those MAPE greater than 1.0, there are large errors relative to the magnitude of the actual data. However, the trend shows that the days (``horizon'') in 40 to 60 keep a relatively low MAPE for the prediction.

\begin{table}[!h]
    \centering
    \caption{}
    \begin{tabularx}{\textwidth}{|c|C|C|C|C|}
        \hline
            & ds                  & yhat       & yhat\_lower & yhat\_upper \\ \hline
        385 & 2024-05-03 20:30:00 & 137.414682 & 82.390317   & 187.754473  \\ \hline
        386 & 2024-05-04 20:30:00 & 156.571010 & 105.134074  & 207.358519  \\ \hline
        387 & 2024-05-05 20:30:00 & 161.983939 & 109.669790  & 210.234837  \\ \hline
        388 & 2024-05-06 20:30:00 & 151.771738 & 99.634336   & 205.890234  \\ \hline
    \end{tabularx}
\end{table}

In Table 5, the meanings of the columns are identical to those in Table 2. It also reveals the last few rows for the results. The estimated range of the ``Quantity Replenished'' is from 137 to 162. Therefore, it is possible to follow the value to control the replenishment quantity, logistics planning and inventory control. The difference between ``yhat\_upper'' and ``yhat\_lower'', gives insight into the variability of the forecast and is helpful for the adjustments of the tactics according to the actual conditions.

\begin{table}[!h]
    \centering
    \caption{}
    \begin{tabularx}{\textwidth}{|C|C|}
        \hline
        Date                & Action needed (units) \\ \hline
        2023-01-01 04:01:00 & 119.64751955133924    \\ \hline
        2023-01-02 20:11:00 & 127.29229423974843    \\ \hline
        2023-01-03 05:17:00 & 118.40997285608174    \\ \hline
        2024-05-05 20:30:00 & 156.57101042096363    \\ \hline
        2024-05-06 20:30:00 & 161.9839387494336     \\ \hline
        2024-05-07 20:30:00 & 151.77173787561628    \\ \hline
    \end{tabularx}
\end{table}

The results from Table 6 are captured by the first and the last 3 rows for the reviewing. The table shows the column Action needed (units) for the ``yhat'' and the Date for the ``ds''. The range of dates is from ``2023-01-01'' to ``2024-05-07''. The model is utilized to make the decisions with a threshold value of 100 units. The values shown in the ``Action needed (units)'' column show the actual predicted quantity for various times and dates. This output is the indicator for making the decisions in the system to plan the demand.

\subsection{Random Forest}
\subsubsection{Code Design}

To look at the code design of Random Forest for the dual-sourcing inventory model, there are some adjustments which are implied based on the code in the lost sales inventory model in the same algorithm. The date columns such as ``Date'', ``Scheduled Delivery Date'' and ``Date of Stock-out'' are converted to datetime and extract the year, month, day, hour and minute components. Some additional features are also simulated, such as ``Supplier Reliability Score'', ``Lead Time'' and ``Cost Difference''. The targeted variable is changed to ``Quantity Replenished''.

To specify the algorithm in the dual-sourcing inventory model, the reordering and decision-making process are applied by ``rf\_best\_model.predict(X\_test\_scaled)''. Creating the dataframe with suppliers, costs and reliability, is used for the decision function to choose the supplier based on the cost and reliability. In the ``place\_order'' function, the minimum order quantity threshold is set to 10 and the corresponding results are the relevant order placed quantity. The loop code is established to decide the order according to the forecast quantity and chosen suppliers.

\subsubsection{Results}

\begin{table}[!h]
    \centering
    \caption{}
    \begin{tabularx}{\textwidth}{|c|C|C|}
        \hline
        MSE          & 0.6762651442865575 & 0.251188744219634   \\ \hline
        MAE          & 0.5227676578348462 & 0.3020580696808377  \\ \hline
        R$^2$  Score & 0.9998460717412805 & 0..9999428256116194 \\ \hline
    \end{tabularx}
\end{table}

The values of MSE and MAE of Random Forest are 0.676 and 0.522 which are relatively low in this prediction in the model. The actual and predicted values are close and the value of MSE indicates a better fit of the model to the data. There is an outstanding anticipated accuracy with the low value of MAE. The $R^2$ Score of the Random Forest, 0.9998 being close to 1 shows that a large portion of the variance occurs and there is a fabulous model performance.

To review the comprehensive results, the Gradient Boosting regressor results are also the indicator for the dual-sourcing inventory model. The value of MSE is 0.251 and it is smaller than the MSE in the Random Forest for better fit of the model to the data. Similarly, the value of MAE in the Gradient Boosting regressor which is 0.302 is smaller than the value of MAE in Random Forest. Thus, it illustrates that the average magnitude of errors in predictions is smaller. The Gradient Boosting regressor gets the 0.999 $R^2$ Score which is slightly larger than the value in Random Forest so it performs a better fit and predictive accuracy.

\begin{table}[!h]
    \centering
    \caption{}
    \begin{tabularx}{\textwidth}{|C|C|c|}
        \hline
        Supplier & Units  & Status                                  \\ \hline
        043      & 0.00   & Order skipped - below minimum threshold \\ \hline
        006      & 77.26  & Order placed                            \\ \hline
        020      & 150.60 & Order placed                            \\ \hline
        006      & 166.68 & Order placed                            \\ \hline
        043      & 69.55  & Order placed                            \\ \hline
        006      & 0.00   & Order skipped - below minimum threshold \\ \hline
    \end{tabularx}
\end{table}

Table 8 is the result of the decision-making regarding order placement and skipping. The units of order are computed by the predicted quantity of replenishment in the Random Forest and Gradient Boosting.

The order is skipped indicating the order quantity is smaller than the threshold. The supplier selection is based on the cost and reliability of the supplier adjusted by a cost difference factor. Therefore, the quantity units represent the placed order exceeding the threshold.

\begin{figure}[!h]
    \centering
    \includegraphics[width=0.8\textwidth]{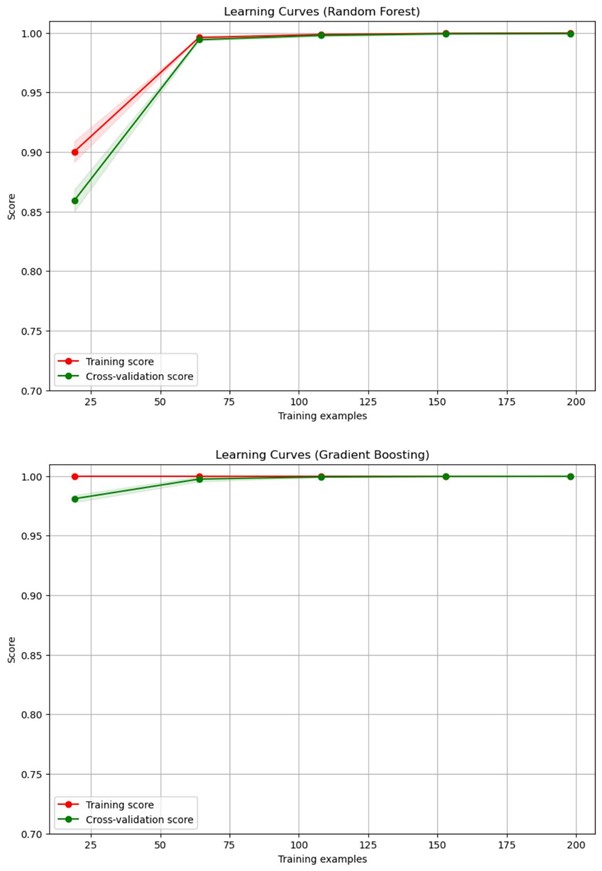}
    \caption*{Figure 20,21}
\end{figure}

Figure 20 shows the training score as a red line, starting at 0.9 on the Random Forest training curve. In training samples between 50 and 75, it continues to rise toward 1.0. As more training examples are added to the model, the score rises steadily and reaches 1.0 in training examples with a score between 100 and 125. As previously stated, it functions well to use the chosen features to reflect the variation in the target variable if the $R^2$ Score is 1. The cross-validation score and continual boosting are represented by the green line. When the number of data increases, the model performs better.

The training score in the Gradient Boosting learning curve is likewise shown by the red line in Figure 21. After it initially achieves a score of 1.0, further data is uploaded. Comparably, the cross-validation score, which begins at a high score, is likewise represented by the green line. The score equals the training score of 1.0 after the model training data is raised.

\begin{figure}[!h]
    \centering
    \includegraphics[width=0.8\textwidth]{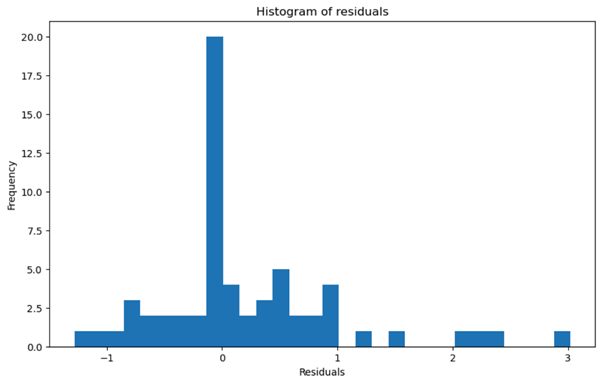}
    \caption*{Figure 22}
\end{figure}

Figure 22 displays the histogram of the residuals along with the residuals frequency. Most residuals stay at zero which means that the model does not overestimate and underestimate the targeted variable (``Quantity Replenished''). Some larger residuals on positive and negative sides are the outliers for the model predictions.

\begin{figure}[!h]
    \centering
    \includegraphics[width=0.8\textwidth]{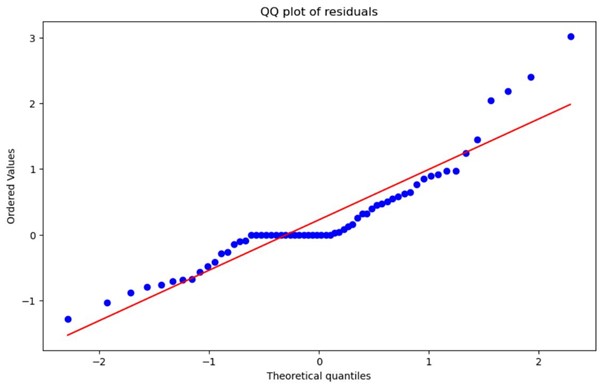}
    \caption*{Figure 23}
\end{figure}

The Q-Q plot of residuals with the x-axis labeled ``Theoretical quantiles'' and the y-axis labeled ``Ordered Values'' is shown in Figure 23 to assess the reliability of data in various theoretical distributions.

Many blue dots are in the middle of the distribution close to the red reference line and the residuals are distributed normally in this range. There are some deviations at the end of the theoretical quantiles which are slightly far from the red reference line.

\subsection{DRL}
\subsubsection{Code Design}

The additional data column ``Scheduled Delivery Date'' and is converted to datetime format. The code handles the missing values for numeric columns and categorical columns (``Supplier ID Encoded'') and adds the new columns relevant to dual-sourcing. This procedure is saved into the new file (``cleaned\_for\_dual\_sourcing\_inventory\_model.csv'').

For the environment construction, the complex action space (``gym.spaces.Discrete(4)'') with dual suppliers is set to be 4. The tracking list for pending orders in suppliers 1 and 2 is separated. For the parameters, the initial inventory levels for the supplier 1 and 2 are set to be 100 respectively and the lead time for those suppliers are 5 days and 10 days. The order cost of supplier 1 is set to be 10 while The order cost of supplier 2 is set to be 15.

The algorithm subtracts the cost of the future order from the current reward and arranges it with supplier 1, supplier 2 and both suppliers. If an order is due today, the completed orders for Suppliers 1 and 2 are excluded and raise inventory by 50 units.

\subsubsection{Results}

\begin{figure}[!h]
    \centering
    \includegraphics[width=0.8\textwidth]{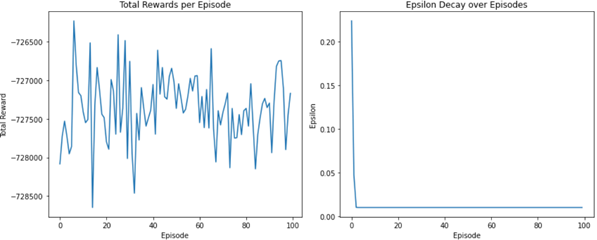}
    \caption*{Figure 24,25}
\end{figure}

In Figure 24, the blue line demonstrates the fluctuation in variability of the agent performance from one episode to the next. The total reward performance fluctuates in the first 40 episodes and becomes more stable from the 40 episodes. The range of the entire total reward performance is from -726000 to - 729000.

For Figure 25, the epsilon decays from the beginning episodes and it remains unchanged at the minimum value of epsilon (= 0. 01).

\subsection{Comparison}

To select the algorithms in the dual-sourcing inventory model in a supermarket context, the inventory is replenished by two different sources suppliers. Supplier and lead time management, cost optimization and inventory level optimization are the potential consideration aspects.

To think about supplier reliability and lead time management, the Time Series algorithm does not effectively capture the variability of the lead times. For the Random Forest (and Gradient Boosting), they are more capable of handling the complex feature interactions and managing the dual-sourcing complexity for forecasting the supplier demand fulfilment. The orders placed with specific suppliers show the capability of decision-making. The DRL potentially excel the supplier reliability and lead times as it is able to learn and adapt the policies according to the reward structure. However, it is based on training with a comprehensive set of features to capture the dynamics of the supplier lead times and reliability. To get fresh goods and respond to the demand alterations, the Random Forest (and Gradient Boosting) are able to process the complex data to manage the lead time appropriately and ensure supplier reliability. For instance, the algorithms use past data from the supplier performance and lead time variability to optimize the ordering times.

To take the cost optimization into consideration, the purpose of the Time Series is not exactly suitable for optimizing the cost. For the Random Forest (and Gradient Boosting), they are able to anticipate the demand accurately and incorporate the cost components in the predictions. Therefore, it is capable of strategic decision-making in the local and overseas suppliers. By incorporating the cost-related factors in the reward function, the actions make saving costs possible to allow the algorithm to learn about cost-effective strategies. Since supermarkets have narrow profit margins, cost-effectiveness is essential. Consequently, the Random Forest (and Gradient Boosting) are the better alternatives to analyze the diverse factors about costs, such as ordering costs, holding costs and the costs of stockout and overstock. It is possible to determine the supplier sources by the unit, holding and transportation costs.

The Time Series is less reliable for inventory level optimization because of shrinking coverage over time. It is more difficult to avoid the stockout. Due to the high accuracy in forecasting the demand, it leads to precise inventory level optimization and diminishes the risk of overstocking and understocking. The DRL algorithm is able to learn the environment repeatedly to adjust the ordering policies. To suit the supermarket context, the algorithm is required to avoid excess inventory as it increases the holding cost and the lost sales. Thus, the Random Forest (and Gradient Boosting) represent the high accuracy to predict the demand to maintain the correct number of stock.

\subsection{Discussion}

\begin{figure}[!h]
    \centering
    \includegraphics[width=0.8\textwidth]{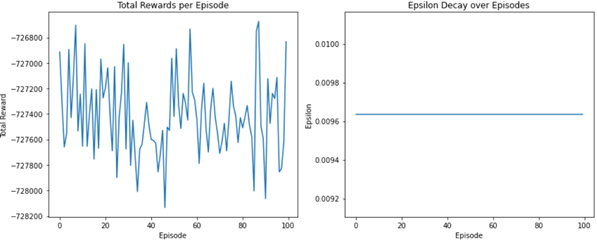}
    \caption*{Figure 26,27}
\end{figure}

To adjust the parameters for getting better results, the two hidden layers with 64 rise to 128 neurons in the network architecture and the dropout rate boosts to 0.3. For the learning rate and the value of ``epsilon\_decay'', they are changed to 0.005 and 0.885 correspondingly.

Despite tuning the parameters, the fluctuation shown in Figure 26 does not emerge as an obvious alteration compared with Figure 24. The range of the total rewards is from -727000 to - 727850. For Figure 27, the epsilon is in a stable circumstance after tuning the parameters. Therefore, the agent is unable to learn in a stable policy to reveal a high variation trend for the total rewards.

\section{Methodologies in Multi-Echelon inventory model}
\subsection{Time Series}
\subsubsection{Code Design}
To construct the algorithm code in this inventory model, the fundamental structure is based on the design in Time Series in the lost sales. However, there are some additional adjustments for Time Series algorithms.

The code selects the specific columns for the dataframe, such as ``Date'' and ``Quantity Replenished''. The columns ``Lead Time'' and ``Market Event E010'' are handled for processing the data and the columns ``Estimated Demand'' and ``Potential Lost Sales'' are imputed the Nan values in the regressors. The dataframe of the prophet is combined by columns ``Date'', ``Quantity Replenished'', ``Potential Lost Sales'', ``Lead Time'', ``Market Event E010'' and ``Estimated Demand''.

For the parameters, the change points prior to scale are changed to 0.05. The constant values for the simplicity are assumed. The initial size is set to be half of the length of the prophet dataframe and the period size is 10\% of the initial size. The horizon is 30\% of the initial size.

\subsubsection{Results}

\begin{figure}[!h]
    \centering
    \includegraphics[width=0.8\textwidth]{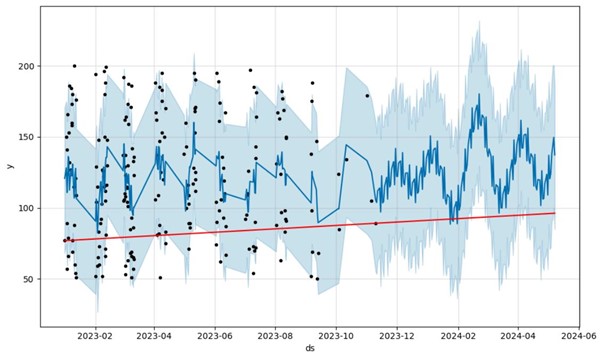}
    \caption*{Figure 28}
\end{figure}

Figure 28 represents the same elements with the x and y-axis, the forecast blue line, the red trend line and the uncertainty interval in other sections. The uncertainty interval shown by the shaded blue area, is analogous before ``2023-10'' with the figure in the lost sales and dual-sourcing inventory models. However, for the date after ``2023-10'', the forecast blue line and the uncertainty interval with the upper bound represent different values. The trend line also changes from 60 to nearly 100.

\begin{figure}[!h]
    \centering
    \includegraphics[width=0.8\textwidth]{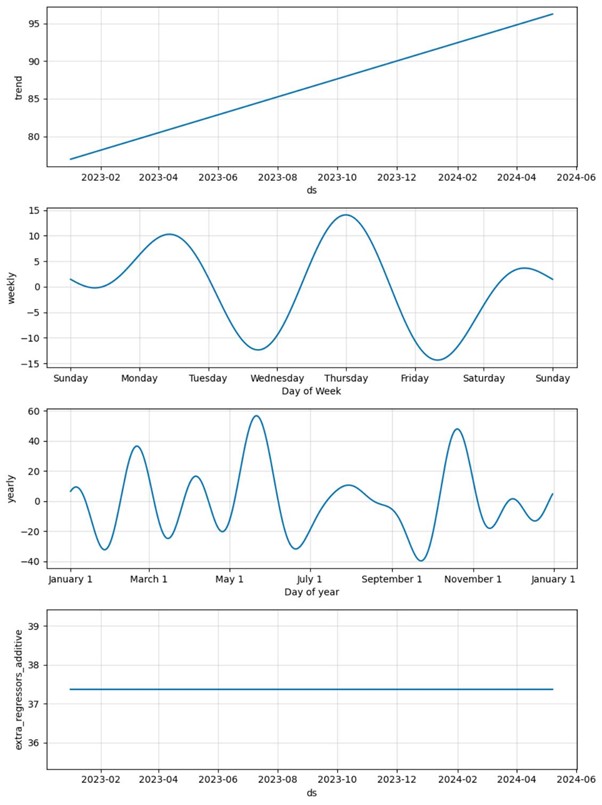}
    \caption*{Figure 29,30,31,32}
\end{figure}

Figure 29 shows an obvious difference in the trend of the ``Quantity Replenished''' over the dates. It illustrates the upward line for the trend and the quantity raises from ``2023-01'' to ``2024-05''. The trend line starts from 75 quantities in ``2023-01'' to 95 in ``2024-05'' approximately.

Figure 30 demonstrates similar patterns with other weekly seasonality figures in other models. The ``Quantity Replenished'' fluctuates within a week. In this graph, the specific restock days and weekly demand cycles are shown.

In Figure 31, the yearly seasonality is displayed to show the ``Quantity Replenished'' fluctuates within a year. There are some obvious peaks and feet in some months.

The influence of the additional regressors is shown in the figure 32. The graph reveals a constant line at the value of 37.2. The regressors do not change the effects over time in the model.

\begin{table}[!h]
    \centering
    \caption{}
    \begin{tabularx}{\textwidth}{|c|C|C|C|C|C|C|C|C|}
        \hline
          & horizon            & mse          & rmse       & mae        & mape      & mdape     & smape     & coverage  \\ \hline
        0 & 2 days    20:32:00 & 3209.36 8825 & 56.6512 91 & 49.4788 55 & 0.42847 7 & 0.31051 3 & 0.40590 8 & 0.59259 3 \\ \hline
        1 & 2 days    22:00:00 & 3136.01 0399 & 56.0000 93 & 48.8090 45 & 0.42649 0 & 0.28936 2 & 0.40311 6 & 0.62963 0 \\ \hline
        2 & 2 days    22:44:00 & 3160.04 7527 & 56.2143 00 & 49.2533 45 & 0.43231 6 & 0.31051 3 & 0.40648 1 & 0.62963 0 \\ \hline
        3 & 3 days    01:17:00 & 3154.04 8372 & 56.1609 15 & 49.1938 46 & 0.44157 1 & 0.33723 1 & 0.40968 2 & 0.62963 0 \\ \hline
        4 & 3 days    03:03:00 & 3214.75 6369 & 56.6988 22 & 49.6902 65 & 0.44985 2 & 0.33723 1 & 0.41362 4 & 0.62963 0 \\ \hline
    \end{tabularx}
\end{table}

Table 9 shows the akin elements with other tables in other models. It shows a few rows in this table. A portion of the initial data set size is utilized for the cross-validation forecast horizon. The time duration in ``Horizon'' column boosts with each row. The performance metrics were evaluated a little more in the future.

The metrics from the table illustrate the unwell performance in the model with the consequential errors and the low coverage. Being unable to capture the underlying patterns in the data is shown by the high values of MAPE and sMAPE.

\begin{table}[!h]
    \centering
    \caption{}
    \begin{tabularx}{\textwidth}{|c|C|C|C|C|}
        \hline
            & ds                  & yhat       & yhat\_lower & yhat\_upper \\ \hline
        385 & 2024-05-03 20:30:00 & 124.190786 & 72.104650   & 175.502172  \\ \hline
        386 & 2024-05-04 20:30:00 & 137.818682 & 90.289098   & 189.819806  \\ \hline
        387 & 2024-05-05 20:30:00 & 143.711743 & 92.273953   & 199.923427  \\ \hline
        388 & 2024-05-06 20:30:00 & 149.658204 & 99.777011   & 200.475138  \\ \hline
        389 & 2024-05-07 20:30:00 & 137.293664 & 84.807996   & 186.279493  \\ \hline
    \end{tabularx}
\end{table}

Table 10 shows a few rows of data for specific dates and times. The replenished quantities are demonstrated in the ``yhat'' column with the dates and times.

The anticipated intervals which are the difference of the ``yhat\_lower'' and ``yhat\_upper'' are relatively wide on each day. It shows that there are some possible uncertainties in the model forecast. The actual ``Quantity Replenished'' is expected to fall within the range of the difference of the ``yhat\_lower'' and ``yhat\_upper''.

\subsection{Random Forest}

\subsubsection{Code Design}

Based on the previous codes in other models, the column ``Replenishment Date'' is converted into datetime format and the targeted variable is ``Quantity Replenished''.

For the additional algorithm ``Gradient Boosting'' besides the Random Forest, its hyperparameters ``n\_estimators'' are chosen from 50 to 200 and ``max\_depth'' is chosen from 2 to 10. The learning rate is continuously uniformly distributed with the lower bound of 0.01 and upper bound of 0.1. For the ``min\_samples\_split'' and ``min\_samples\_leaf'', they are chosen from 2 and 1 to 20 respectively.

\subsubsection{Results}

\begin{table}[!h]
    \centering
    \caption{}
    \begin{tabularx}{\textwidth}{|c|C|C|}
        \hline
                     & Random Forest      & Gradient Boosting  \\ \hline
        MSE          & 1341.3523259133704 & 1475.8167850784525 \\ \hline
        MAE          & 27.35344098928057  & 28.76369257899875  \\ \hline
        R$^2$  Score & 0.6946877573070183 & 0.6640815960493969 \\ \hline
    \end{tabularx}
\end{table}

In Table 11, for the Random Forest performance metrics, a modest level of results is shown to forecast ``Quantity Replenished''. With the MSE (1341.35) and the MAE (27.35) in Random Forest, these two metrics reveal that the average errors for the forecast model and the performance are not well enough. An excellent fit between the observed data and the model is shown by the $R^2$ Score of around 0.695, which accounts for more than 70\% of the variation in the replenishment data.

Additionally, the MSE (1475.81) and MAE (28.76) are relatively high to lead to a low accuracy. The $R^2$ Score is around 0.664 and it explains that there are 66\% of variation in the data.

The predictive power of both models for inventory replenishment is modest, capturing a considerable portion of the variability in the data but yet allowing room for unknown causes.

\begin{figure}[!h]
    \centering
    \includegraphics[width=0.8\textwidth]{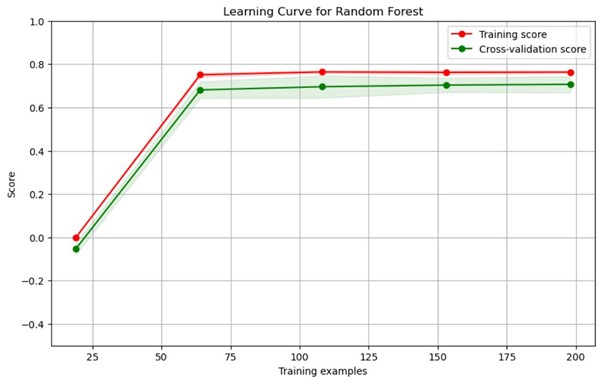}
    \caption*{Figure 33}
\end{figure}

The training score begins at 0, enhances to nearly 0.8 between 75 and 100 training examples and stays at a falt circumstance with more training data. Since the Random Forest model is capable of capturing the patterns in the data.

The cross-validation score, starts with a negative score and boosts to nearly 0.7 scores in between 75 and 100 training examples. Then, it becomes more falt to add more training data.

Particularly when more data is utilized, the difference between the cross-validation score and the training score is not that great. The model does not exhibit considerable overfitting and is well-generalizing.

\begin{figure}[!h]
    \centering
    \includegraphics[width=0.8\textwidth]{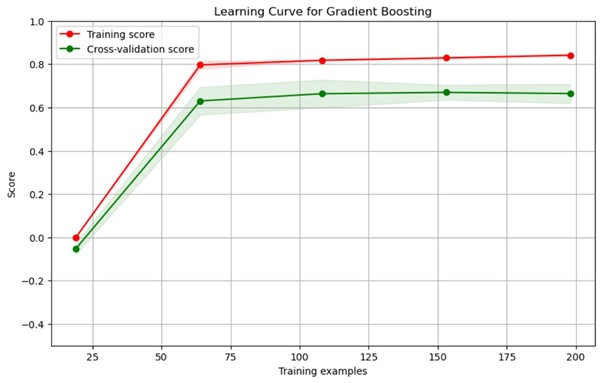}
    \caption*{Figure 34}
\end{figure}

The model performs well with training data and becomes better with additional data on the cross-validation set; nevertheless, there appears to be a plateau effect, suggesting that the gains from adding more training data may reach a limit. The high and consistent results across various training set sizes suggest that the model is tailored to the complexity of the data.

\begin{figure}[!h]
    \centering
    \includegraphics[width=0.8\textwidth]{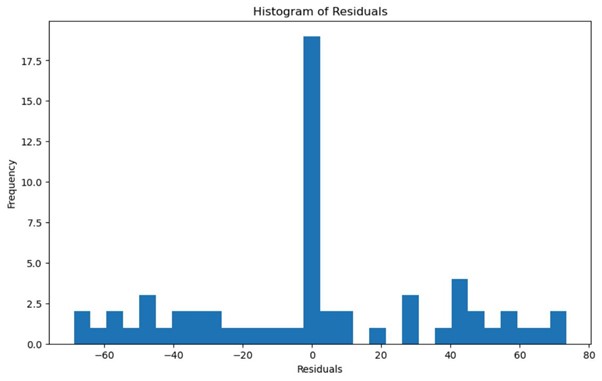}
    \caption*{Figure 35}
\end{figure}

The distribution of residuals between the values forecasted by the Random Forest and the observed values of the target variable is demonstrated in the histogram.

A significant amount of residuals are centered around zero, indicating that the model was quite accurate for a large number of predictions.

The imbalance is seen in the residuals, with more positive than negative residuals. This implies that rather than overestimating the actual values, the model most of the time underestimated them.

\begin{figure}[!h]
    \centering
    \includegraphics[width=0.8\textwidth]{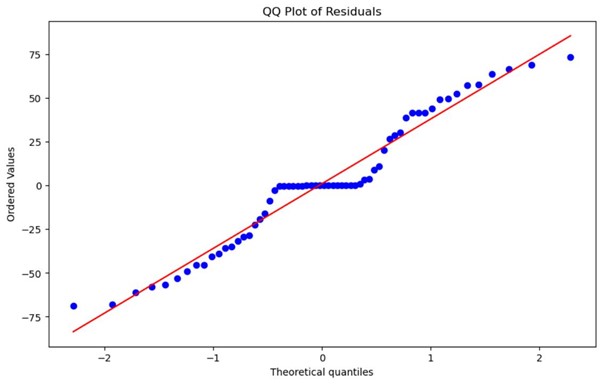}
    \caption*{Figure 36}
\end{figure}

Figure 36 is a quantile-quantile plot for the residuals from the Random Forest model predictions.

Most points follow the red line, specifically in the middle of the red line. It shows that the residuals are nearly normally distributed in the range.

But the points in the beginning and ending have heavier tails than the normal distribution.

\subsection{DRL}
\subsubsection{Code Design}
The action space is set to 4 and the parameter ``order cost'' is increased to 20. For the holding cost and stockout cost, they are altered to 2 and 100. The transfer cost which is 15 is the cost of moving the items between the echelons.

In the ``def step(self, action)'' function, if the action is 0, the days until replenishment are simulated to 5. While the action is 1, the days until replenishment are simulated to 3. For actions 0 and 1, the value of the reward is decreased by the total cost of an order, calculated by multiplying the order cost and the order quantity. If the action becomes 2, it means that the stock is transferred from echelon 1 to 2. The cost of the transference quantity is 20 and the value of the reward is diminished by the transfer cost and the quantity sold is increased by the amount of transference quantity simulating a sales boost from transferring stock. The discount rate is set to be 0.95 in this model. The layers are connected with 128 units and use the ReLU activation function.

\subsubsection{Results}

\begin{figure}[!h]
    \centering
    \includegraphics[width=0.8\textwidth]{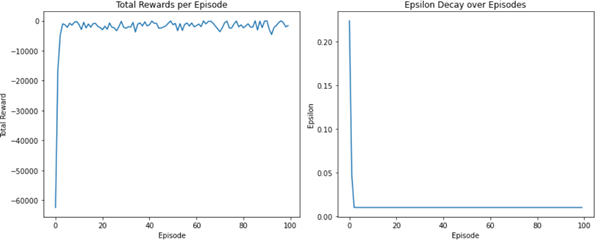}
    \caption*{Figure 37,38}
\end{figure}

The reward in Figure 37 shows a relatively stable after the initial variation. It indicates that the agent's performance of the reward reaches the plateau. The stable policy for the environment is learnt by the agent. The reward nearly stays at 0 after the initial episode.

In Figure 38, epsilon begins to decrease at the beginning of the episodes and remains constant at the lowest value of epsilon (=0.01).

\subsection{Comparison}

To compare the supermarket context in the multi-echelon inventory model, supplier reliability and lead time management, cost optimization and inventory level optimization are also considerable aspects. In the multi-echelon inventory model, there are different levels of the supply chain, such as central warehouses, regional distribution centers and individual stores.

For supplier reliability and lead time management, supermarkets rely on the supplier's network with various lead times and reliability. If there are some delays for the products, the entire supply chain would be affected. Therefore, anticipating arrival times and punctuality for the products contribute to efficient management in this aspect. In the Time Series algorithm, the forecasting horizon boosts leading to longer lead times for the products. Its high RMSE indicates that it is not accurate to forecast the timing of the inventory needs and the delivery schedules. The Random Forest (and Gradient Boosting) algorithms are capable of handling the supplier performance data and also analysing the historical data patterns of the reliability and lead times. Thus, they could provide more reliable predictions for the delivery schedules. The DRL agent adjusts the ordering policies by studying the action sequences and it considers the supplier reliability and lead times. The ``Total Rewards per Episode'' illustrates a stable trend and indicates the consistent performance of the agent. The higher rewards the agent performs, the better handling results lead times and supplier reliability have. Based on this circumstance, the Random Forest and Gradient Boosting are able to handle multiple variables and more complex data.

Optimizing the cost with different supply chain stages and various costs of products is the main goal in the supermarket context in the multi-echelon inventory model. The time Series algorithm does not capture all the cost factors adequately leading to a less accurate demand prediction. Therefore, it is possible to contribute to the high holding costs and high shipping costs. The Random Forest and Gradient Boosting are capable of incorporating a wide range of influential factors and identifying the patterns. The results of these algorithms indicate that they are able to anticipate the demand precisely and manage the inventory to avoid emerging overstocking and immediate orders. In the DRL environment construction, the order costs, holding costs and stockout costs are set in that section. Thus, the agent is capable of learning from different costs. Despite the stable line in the graph, an upward and increasing trend in the total rewards is expected for cost optimization. Consequently, the Random Forest and Gradient Boosting are more suitable in the aspect of utilizing detailed feature engineering to contain the cost factors in the forecast models.

To have the optimization inventory level in all echelons, reaching the customer demand and decreasing the holding costs. The limited coverage by the Time Series model is not reliable because of the high volatility which leads to poor inventory levels. For the Random Forest and Gradient Boosting, the metrics such as a lower MSE and higher $R^2$ scores from these algorithms illustrate that the model is able to forecast the demand more accurately and it makes better inventory levels. The performance shown in the above DLR graphs demonstrates that a steady policy for the inventory levels. But the further inventory optimization by the agent is uncertain. Therefore, Random Forest and Gradient Boosting models predict the accuracy of inventory level optimization.

\subsection{Discussion}

\begin{figure}[!h]
    \centering
    \includegraphics[width=0.8\textwidth]{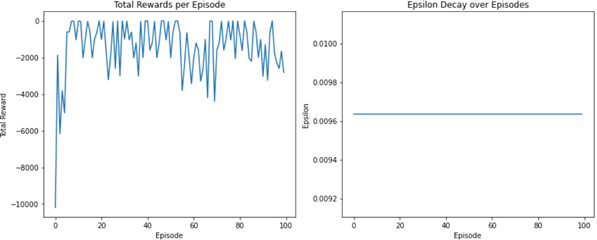}
    \caption*{Figure 39,40}
\end{figure}

To get the alternative of the performance, the ``epsilon\_decay'', learning rate and dropout rate are altered to 0.885, 0.005, and 0.3 respectively.

After tuning the parameters in this model, the total rewards in Figure 39 show more fluctuation but the rewards plot in Figure 37 represents smooth and consistent relatively. The training results show a stable learning process before tuning the parameters with the agent achieving a relatively consistent reward after initial episodes. The variance is also increased due to a higher exploration rate. For the ``Epsilon Decay over Episodes'' in Figure 40, it shows a stable epsilon value and the agent explores less as training progresses.

\section{Conclusion}
This report compared three methodologies—Time Series (TS), Random Forest (RF), and Deep Reinforcement Learning (DRL)—for managing supermarket inventories. The findings reveal each method's strengths in tackling common retail challenges such as lost sales, dual-sourcing, and coordinating multi-echelon inventory models.

In the context of the lost sales inventory model, all three methodologies demonstrate their ability to minimize lost sales and optimize inventory levels, with DRL exhibiting the most promising results. For the dual-sourcing inventory model, TS, RF, and DRL each effectively optimize the balance between regular and express sources, leading to reduced costs and improved service levels. In the multi-echelon inventory model, all three approaches showcase their ability to coordinate inventory decisions across multiple levels of the supply chain, with DRL demonstrating the highest efficiency.

The results of this study contribute to the existing body of knowledge on data-driven inventory management and provide valuable insights for retailers. Theoretically, this research contributes empirical data supporting the efficacy of TS, RF, and DRL in complex inventory scenarios. Each method brings unique strengths and limitations, varying by retail setting and inventory model.

From a practical standpoint, supermarkets stand to gain significantly by adopting these data-driven systems. Such tools can boost operational efficiency, cut costs, and improve customer satisfaction. This study serves as a guide for retail managers to choose the best-suited method for their specific needs.

This research evaluates various methodologies and their real-world relevance to inventory management, bridging theoretical knowledge and practical implementation. Key contributions include:
\begin{enumerate}
    \item   Enhancing understanding of data-driven inventory management approaches and their specific advantages and challenges.
    \item   Offering a framework to assess and select the best methods based on the unique aspects of different retail environments.
    \item   Demonstrating how these methodologies can refine inventory management in the retail sector.
\end{enumerate}

While insightful, this study has limitations that future research could address. Expanding data sources to include external factors like economic trends and consumer preferences could enhance algorithm accuracy. A longitudinal analysis would also shed light on these methodologies' long-term efficacy.

Future studies might explore a hybrid system combining TS, RF, and DRL strengths. Such a system must be developed closely with retail stakeholders to meet specific industry needs.

In summary, this research underscores the transformative potential of TS, RF, and DRL in retail inventory management. By leveraging advanced analytics and machine learning, supermarkets can refine their inventory strategies, reduce costs, and boost customer satisfaction. These findings lay a robust groundwork for further innovation in data-driven inventory management, paving the way for a more efficient and customer-focused retail future. As the industry evolves, embracing these data-driven approaches will be crucial for staying competitive and addressing the dynamic needs of consumers.

\bibliographystyle{plainnat}
\bibliography{reference}



\end{document}